\documentclass[runningheads,a4paper]{llncs}
\usepackage{amsmath, amssymb}
\usepackage{graphicx}
\usepackage{amsfonts}
\usepackage{xcolor}
\usepackage{caption}
\usepackage{subcaption}
\usepackage{algorithm}
\usepackage{algpseudocode}
\usepackage[inline]{enumitem}
\usepackage{bm}
\usepackage{booktabs}
\usepackage{hyperref} 

\begin{document}
\title{Bayesian Optimization for Function Compositions with Applications to Dynamic Pricing}
% \title{Pricing/Demand Curve Optimisation using Novel Bayesian Optimisation Techniques}
%

\titlerunning{BO for Function Compositions}

\author{Kunal Jain\inst{1}, Prabuchandran K. J.\inst{2} \and Tejas Bodas\inst{1}}

\authorrunning{K. Jain et al.}

\institute{International Institute of Information Technology, Hyderabad \\ \email{\{kunal.jain@research.,tejas.bodas@\}iiit.ac.in} \and Indian Institute of Technology, Dharwad \\ \email{prabukj@iitdh.ac.in}}

\footnotetext[2]{Prabuchandran K.J. was supported by the Science and Engineering Board (SERB), Department of Science and Technology, Government of India for the startup research grant ‘SRG/2021/000048’.}
\maketitle              % typeset the header of the contribution
\begin{abstract}
Bayesian Optimization (BO) is used to find the global optima of black box functions. In this work, we propose a practical BO method of function compositions where the form of the composition is known but the constituent functions are expensive to evaluate. By assuming an independent Gaussian process (GP) model for each of the constituent black-box function, we propose Expected Improvement (EI) and Upper Confidence Bound (UCB) based BO algorithms and demonstrate their ability to outperform not just vanilla BO but also the current state-of-art algorithms. We demonstrate a novel application of the proposed methods to dynamic pricing in revenue management when the underlying demand function is expensive to evaluate.

\end{abstract}

\section{Introduction}
Bayesian Optimization (BO) is a popular technique for optimizing expensive-to-evaluate black-box functions. Such a function might correspond to the case where evaluating it can take up to hours or days, which for example, is the case in re-training massive deep learning models with new hyper-parameters~\cite{wu2019hyperparameter}. In some cases, functions can be financially costly to evaluate, such as drug testing ~\cite{pyzer2018bayesian} or revenue maximization~\cite{phillips2021pricing}. In such black-box optimization problems, one often has a fixed budget on the total number of function evaluations that can be performed. For example, one typically has a budget on the computational capacity spent in the hyper-parameter tuning of neural networks~\cite{pmlr-v28-bergstra13,https://doi.org/10.48550/arxiv.1206.2944}. In such cases, BO proves to be very effective in providing a resource-conserving iterative procedure to query the objective function and identify the global optima~\cite{Perrotolo2018ATF}.

The key idea in BO is to use a surrogate Gaussian Process (GP) model~\cite{Rasmussen2005} for the black-box function, which is updated as and when multiple function evaluations are performed. To identify the location (in the domain) of the following query point, an acquisition function (a function of the surrogate model) is designed and optimized. The design of the acquisition function depends not only on the application in mind but also on the trade-off between exploration and exploitation that comes naturally with such sequential optimization problems. Some popular acquisition functions used in the literature are the Expected Improvement(EI), Probability of Improvement(PI), Knowledge Gradient(KG) and Upper Confidence Bound(UCB) \cite{https://doi.org/10.48550/arxiv.1012.2599,https://doi.org/10.48550/arxiv.1807.02811,Candelieri2021}. 
%Work is also being done to find fast and accurate methods to optimize these functions~\cite{https://doi.org/10.48550/arxiv.2111.04930}.

In this work, we consider Bayesian optimization of composite functions of the form $g(x) = h(f_1(x), f_2(x), \ldots f_M(x))$ where functions $f_i$ are expensive black-box functions while $h$ is known and easy to compute. More specifically, we are interested in maximizing $g$ and identifying its optima. A vanilla BO approach can be applied to this problem, ignoring the composite structure of the function~\cite{Candelieri2021,https://doi.org/10.48550/arxiv.1807.02811}. In this approach, one would build a GP posterior model over the function $g$ based on previous evaluations of $g(x)$ and then select the next query point using a suitable acquisition function computed over the posterior. However, such a vanilla BO approach ignores the available information about the composite nature of the functions, which we show can easily be improved upon. In this work, we model each constituent function $f_i$ using an independent GP model and build acquisition functions that use the known structure of the composition. Our algorithms outperform the vanilla BO method as well as the  state-of-art method ~\cite{https://doi.org/10.48550/arxiv.1906.01537} in all test cases and practical applications that we consider. Our algorithms are also more practical and less computationally intensive than the methods proposed in~\cite{https://doi.org/10.48550/arxiv.1906.01537}.

Note that function compositions arise naturally in the real world. One such example is the revenue maximization problem based on the composition of price and demand function. Another example could be the optimization of the F1 score in classification problems that can be seen as a composition of precision and recall metrics~\cite{8647714}. A key novelty of our work lies in the application of our BO algorithms to dynamic pricing problems in revenue management. To the best of our knowledge, ours is the first work to perform dynamic pricing for revenue optimization using Bayesian optimization methods.  
%In optimization functions where the final metric is evaluated using a known composite of it's output, leveraging the composite function is a powerful method to converge to the optimum faster. \textcolor{red}{elaborate}
\subsection{Related Work}
Optimization of function compositions has been studied under various constraints such as convexity, inexpensive evaluations and derivative information ~\cite{barber2016mocca,NIPS2016_645098b0,https://doi.org/10.48550/arxiv.1605.00125,shapiro2003}. The scope of these work are somewhat restrictive and differ from our key assumption that the constituent functions in the composition are black-box functions and are relatively expensive to evaluate. Our work is closely related to Astudillo and Frazier~\cite{https://doi.org/10.48550/arxiv.1906.01537} who optimize black-box function compositions of the form $g(x) = h(f(x))$ using Bayesian Optimization. In this work, the constituent function $f(x)$ is expensive to evaluate and is modelled as a Multi-Ouput GP. This work was further improved by Maddox et al.~\cite{maddox2021bayesian} using Higher Order Gaussian Process (HOGP). These work assume that the member functions in the composition are correlated and dedicates a significant amount of computational power to capturing these correlations. They propose an EI-based acquisition function for estimating the value of $g$ using a MOGP over the functions $f$. The calculation of this acquisition function requires them to compute the inverse of the lower Cholesky factor of the covariance matrix, which is a computationally expensive task (runtime increases with order $\mathcal{O}(N^3)$ where $N$ is the size of the covariance matrix), especially when optimizing high dimensional problems.

Our work differs from them in that we consider a composition of multiple constituent functions with single output and assume an independent GP model for each such constituent. This results in significantly lesser computational requirements and faster iterations. Our work focuses on practical deployments of the technique, as showcased in Section~\ref{sec:experiments}.  We also propose a UCB-based algorithm where the problem of matrix inversion does not arise. The UCB-based algorithm allows the user to trade-off between exploration and exploitation during iterations, making it more practical for our use cases. Finally, our key contribution lies in applying the proposed methods to dynamic pricing problems, a brief background of which is discussed in the next subsection.

\subsection{Dynamic Pricing and Learning}
Dynamic pricing is a phenomenon where the price for any commodity or good is changed to optimize the revenue collected. Consider the scenario of a retailer with a finite inventory, finite time horizon and a known probabilistic model for the demand. On formulating this as a Markov decision problem, it is easy to see that a revenue optimal pricing policy would be non-stationary, resulting in different optimal prices for the same inventory level at different time horizons. In this case, the dynamic nature of pricing is a by-product of finite inventory and horizon effects. See \cite{phillips2021pricing} for more details.

Now consider a second scenario of a retailer with an infinite horizon and infinite inventory, trying to find the optimal price for his product. Assuming that the underlying probabilistic demand model is unknown to the retailer, this becomes a simultaneous demand learning and price optimization problem. To learn the underlying demand function, the retailer is required to probe or explore the demand for the product at various prices, and use the information gathered to converges to an optimal price over time. Clearly, in this setting, uncertainty in the demand process naturally leads to exploration in the price space, resulting in the dynamic nature of the pricing policy. See \cite{den2015dynamic} for more details. This second scenario (black-box demand function) is of particular interest to us, and we apply our BO algorithms in this setting, something that has not been done before.

Dynamic pricing with learning is a traditional research topic in Operations Research with a long history (see \cite{den2015dynamic2} for a historical perspective). Lobo and Boyd~\cite{e21070651} introduced an exploration-exploitation framework for the demand pricing problem, which balances the need for demand learning with revenue maximization. The problem has been studied under various conditions, such as limited~\cite{10.1145/2559152} and unlimited inventory~\cite{harrison2012bayesian}, customer negotiations~\cite{kuo2011dynamic}, monopoly~\cite{crapis2017monopoly,chen1992dynamic}, limited price queries~\cite{cheung2017dynamic} etc.
One recurrent theme in these work is to assume a parametric form for the demand function in terms of price and other exogenous variables. Reinforcement learning (RL) methods are then used to simultaneously learn the unknown parameters of the demand function and set prices that have low regret.  See, for example,  Broder and Rusmevichientong~\cite{10.2307/23260288} where linear, polynomial and logic demand function models have been assumed. To the best of our understanding, these assumptions on the demand function are rather over-simplified and are typically made for technical convenience. Further, an RL method suited for a particular demand model (say linear demand) may not work when the ground truth model for the demand is different (say, logit model). There have been recent models which try to avoid this issue by modelling the demand as a parameterized random variable (here price is a parameter) but end up making similar convenient  assumptions on the parametric form of the mean or variance of the demand, see \cite{den2014simultaneously} and references therein.

To keep the demand function free from any specific parametric form, in this work, we assume that it is a black-box expensive to evaluate function and instead model it using a Gaussian process. The revenue function can be expressed as a composition of the price and the demand and this allows us to apply our proposed methods (of Bayesian optimization for function composition) to the dynamic pricing and learning problem.

\subsection{Contributions and Organization}
The following are the key contributions of our work:
\begin{enumerate}
    \item We propose novel acquisition functions cEI and cUCB for Bayesian Optimization for function compositions. These acquisition functions are based on EI and UCB acquisition functions for vanilla BO and are less compute intensive and faster to run through each iteration as compared to other state of the art algorithms for function composition.
    \item We assume independent GPs to model the constituent functions and this allows for possible parallelization of the posterior update step.
    \item As a key contribution of this work, we propose to use BO based dynamic pricing algorithms to optimize the revenue. To the best of our knowledge, we are the first, to use BO for learning the optimal price when the demand functions are expensive to evaluate or are black-box in nature.
    \item We consider various revenue maximization scenarios, obtain the revenue function as a composition of price and demand and illustrate the utility of our algorithms in each of these setting.
\end{enumerate}

The following is the organization of the rest of the paper. Section~\ref{sec:problem_description} formally describes the problem statement and Section~\ref{sec:our_approach} describes our proposed algorithms; Section~\ref{sec:experiments} details our experimental results ; and finally, we conclude in Section~\ref{sec:future_works}.
\section{Problem Description}\label{sec:problem_description}
We begin by describing the problem of BO for composite functions in subsection \ref{sec:desc_BO}. In subsection \ref{sec:demand_model} we describe the dynamic pricing problem and model the revenue function as a function composition to which BO methods for composite functions can be applied. 
\subsection{BO for Function Composition}\label{sec:desc_BO}
We consider the problem of optimizing $g(\textbf{x}) = h(f_1(\textbf{x}), f_2(\textbf{x}), \ldots , f_M(\textbf{x}))$ where $g : \mathcal{X} \rightarrow \mathrm{R}$, $f_i : \mathcal{X} \rightarrow \mathrm{R}$, $h:\mathrm{R}^{M} \rightarrow \mathrm{R}$ and $\mathcal{X} \subseteq \mathrm{R}^d$. We assume each $f_i$ is a black-box expensive-to-evaluate continuous function while $h$ is known and cheap to evaluate given the values of $f_i$. The optimization problem that we consider is 
\begin{equation}\label{eq:problemdesc}
  \max_{\textbf{x} \in \mathcal{X}} ~h(f_1(\textbf{x}), f_2(\textbf{x}), \ldots, f_M(\textbf{x})). 
\end{equation}
We want to solve Problem~\ref{eq:problemdesc} in an iterative manner where in the $n^{\text{th}}$ iteration, we can use the previous observations $\{\textbf{x}_i, f_1(\textbf{x}_i), \ldots, f_M(\textbf{x}_i\})\}^{n-1}_{i=1}$ to request a new observation $\{\textbf{x}_n, f_1(\textbf{x}_1), \ldots, f_M(\textbf{x}_n)\}$.

A vanilla BO algorithm applied to this problem would first assume a prior GP model on $g$, denoted by $\mathcal{GP}(\mu(\cdot), K(\cdot, \cdot))$ where $\mu$ and $K$ denote the mean and covariance function of the prior model. Given some function evaluations, an updated posterior GP model is obtained. A suitable acquisition function, such as EI or PI can be used, to identify the next query point. For example, in the $n+1^{\text{th}}$ update round, one would first use the $n$ available observations $(g(\textbf{x}_1), g(\textbf{x}_2), \ldots, g(\textbf{x}_n))$ to update the GP model to  $\mathcal{GP}(\mu^{(n)}(\cdot), K^{(n)}(\cdot, \cdot))$   where $\mu^{(n)}(\cdot)$ is the posterior mean function and $K^{(n)}(\cdot, \cdot)$ is the posterior covariance function, see \cite{Rasmussen2005} for more details. The acquisition function then uses this posterior model to identify the next query location $\textbf{x}_{n+1}$. In doing so, vanilla BO ignores the values of the member functions in the composition $h$.

BO for composite function, on the other hand, takes advantage of the available information about $h$, and its easy-to-compute nature. Astudillo and Frazier~\cite{https://doi.org/10.48550/arxiv.1906.01537} model the constituent functions of the composition by a single multi-output function $\textbf{f}(\textbf{x}) = (f_1(\textbf{x}), \ldots, f_M(\textbf{x}))$ and then model the uncertainty in $\textbf{f}(\textbf{x})$ using a multi-output Gaussian process to optimize $h({\textbf{f}(\textbf{x})})$. Since the prior over $f$ is modelled as a MOGP, the proposed method tries to capture the correlations between different components of the multi-output function $\textbf{f}(\textbf{x})$. Note that the proposed EI and PI-based acquisition functions are required to be computed using Monte Carlo sampling. Furthermore, a sample from the posterior distribution is obtained by first sampling an $n$ variate normal distribution, then scaling it by the lower Cholesky factor and then centering it with the mean of the posterior GP. Two problems arise due to this: \begin{enumerate*} 
    \item Such simulation based averaging approach increases the time complexity of the procedure  linearly with the number of samples taken for averaging and
    \item calculation of the lower Cholesky factor increases the function's time complexity cubically with the number of data points. 
\end{enumerate*}
These factors render the algorithm unsuitable, particularly for problems with large number of member functions or for problems with large dimensions.

% \textbf{Expand on this, why sampling and inversion is needed, describe their acquisiton in 1-2 lines, enumerate drawbacks of astuldo work, 1) computationally expensive 2) inversion 3) 1&2 make it slow and not suitable for large dimension problems  }

To alleviate these problems, in this work, we model the constituent functions using independent GPs. This modelling approach allows us to train GPs for each output independently and hence the posterior GP update can be parallelized. We propose two acquisition functions, cEI which is based on the EI algorithm and cUCB, which is based on the GP-UCB algorithm \cite{Srinivas_2012}. Our cEI acquisition function is similar in spirit to the EI-CF acquisition function of \cite{https://doi.org/10.48550/arxiv.1906.01537} but is less computationally intensive owing to the independent GP model. Since we have independent one dimensional GP model for each constituent function, sampling points from the posterior GP does not require computing the Cholesky factor (and hence matrix inversion), something that is needed in the case of high-dimensional GP's of \cite{https://doi.org/10.48550/arxiv.1906.01537}.  
This greatly reduces the complexity of the MC sampling steps of our algorithm (see section \ref{sec:our_approach} for more details). However, the cEI acquisition function still suffers from the drawback of requiring Monte Carlo averaging. 
To alleviate this problem, we propose a UCB based acquisition function that uses the current mean plus scaled variance of the posterior GP at a point as a surrogate for the constituent function at that point. As shown by Srinivas et al.~\cite{Srinivas_2012}, while the mean term in the surrogate guides exploitation, it is the variance of the posterior GP at a point that allows for suitable exploration. The scaling of the variance term is controlled in such a way that it balances the trade off between exploration and exploitation. In Section~\ref{sec:experiments}, we illustrate the utility of our method, first for standard test functions and then as an application to dynamic pricing problem. Our algorithms, especially the cUCB one, outperforms not only vanilla BO but also those proposed in Astudillo and Frazier~\cite{https://doi.org/10.48550/arxiv.1906.01537}.

\subsection{Bayesian Optimization for dynamic pricing}\label{sec:demand_model}
We consider Bayesian Optimization for two types of revenue optimization problems. The first problem optimizes the revenue per customer where customers are characterized by their willingness-to-pay distribution (which is unknown). In the second problem, we assume a parametric demand model (the functional form is assumed to be unknown) and optimizes the associated revenue. 

In the first model, we assume that an arriving customer has an associated random variable, $V$, with complimentary cumulative distribution function $\bar{F}$, indicating its maximum willingness to pay for the item sold. For an item on offer at a price $p$, an arriving customer purchases it with the probability 
\begin{equation}\label{eq:fp}
    d(p):= \bar{F}(p) = \text{Pr}\{V \geq p\}  .
\end{equation}
In this case, when the product is on offer at a price $p$, the revenue per customer $r(p)$ is given by $r(p) = p \bar{F}(p)$. The revenue function is a composition of the price and demand or purchase probability and we assume that the distribution of the purchase probability i.e., $F$ is not known and also expensive to estimate. One could perform a vanilla BO algorithm by having a GP model on $r(p)$ itself. However to exploit the known nature of the revenue function, we will apply our function composition method by instead having a GP on  $\bar{F}(p)$ and demonstrate its superiority over vanilla BO.  

In the second model, we assume that the true demand $d(p)$ for a commodity at price $p$ has a functional form. This forms the ground truth model that governs the demand, but we assume that the functional form for this demand is not known to the manager optimizing the revenue. In our experiments, we assume linear, logit, Booth and Matyas functional forms for the demand (see section \ref{sec:experiments} for more details.)
Along similar lines, one could build more sophisticated demand models to account for external factors (such as supply chain issues, customer demographics or inventory variables), something that we leave for future explorations. 
%in the model by modelling demand as a parametric function of price
%\begin{equation}
 %   d(p, \boldsymbol{z}):= \bar{F}_{\boldsymbol{z}}(p) = \text{Pr}\{V \geq p\}
%\end{equation}
%where $z$ is the function parameter over $\bar{F}$ which can evolve over time with changing conditions.

Note that we make some simplifying assumptions about the retail environment in these two models and our experiments. We assume a non-competitive monopolistic market with an unlimited supply of the product and no marginal cost of production. However, these assumptions can easily be relaxed by changing the ground truth demand model appropriately, which are used in the experiments to reflect these aspects. The fact that we use a GP model as a surrogate for the unknown demand model offers it the ability to model a diverse class of demand functions under diverse problem settings. We do not discuss these aspects further but focus on the following simple yet meaningful experimental examples that one typically encounters in revenue management problems.  

In the following, $\textbf{p}$ denotes the price vector:
\begin{enumerate}
    \item \textbf{Independent demand model:} A retailer supplies its product to two different regions whose customer markets behave independently from each other. Thus, the same product has independent and different demand functions (and hence different optimal prices) in different geographical regions and under such black-box demand models for the two regions $(d_1, d_2)$. The retailer is interested in finding the optimal prices, leading to the optimization of the following function: $g(\textbf{p}) = p_1d_1(p_1) + p_2d_2(p_2)$. 
    \item \textbf{Correlated demand model:} Assume that a retailer supplies two products at prices $p_1$ and $p_2$ and the demand for the two products is correlated and influenced by the price for the other product. 
    %product to two regions, but their customer market does not behave independently from each other, leading to the objective function 
    Such a scenario can be modelled by a revenue function of the form $g(\textbf{p}) = p_1d_1(\textbf{p}) + p_2d_2(\textbf{p})$. Consider the  example where the prices of business and economy class tickets can influence the demand in each segment. Similarly, the demand for a particular dish in a fast food chain might be influenced by the prices for other dishes.
    \item \textbf{Identical price model: } In this case, the retailer is compliant with having a uniform price across locations. However, the demand function across different locations could be independent at each of these locations, leading to the following objective function: $g(p) = pd_1(p) + pd_2(p)$. This scenario can be used to model different demand functions for different population segments in their age, gender, socio-economic background, etc. 
\end{enumerate}
%Note that the first scenario is a special case of the second scenario where function $d_i$ is only dependent on $p_i$.
\section{Proposed Method}\label{sec:our_approach}
% \subsection{Overview}
% \textbf{Change this according to section 2 changes }
As discussed in Section~\ref{sec:problem_description}, we propose that instead of having a single GP model over $g$, we have $M$ different GP models over each constituent function in the composition. Each prior GP model will be updated using GP regression whenever the observations of the constituent functions are available. A suitably designed acquisition function would then try to find the optimal point when the constituent functions should all be evaluated at, in the next iteration. For ease of notation, we use the shorthand $h(\{f_i(\textbf{x})\})$ to denote $h(f_1(\textbf{x}), \ldots, f_M(\textbf{x}))$ in the subsequent sections.

\subsection{Statistical Model and GP regression}
Let $f_i^{1:n},~ i\in\{1,2,\ldots,M\}$ denote the function evaluations of the member functions at locations $\{\textbf{x}_1,\textbf{x}_2, \ldots, \textbf{x}_n\}$ denotes as $\textbf{x}^{1:n}$. In the input space $\mathcal{X} \subset \mathbb{R}^d$, let $\mathcal{GP}(\mu^{(n)}_i, k^{(n)}_i)$ be the posterior GP over the function $f_i$ where $\mu^{(n)}_i : \mathcal{X} \rightarrow \mathbb{R}$ is the posterior mean function, $k^{(n)}_i : \mathcal{X} \times \mathcal{X} \rightarrow \mathcal{R}$ is the positive semi-definite covariance function and the variance of the function is denoted by $\sigma_i^{(n)}(\textbf{x})$. The superscript $n$ is used to denote the fact that the posterior update accounts for $n$ function evaluations made till now.
For each such GP, the underlying prior is a combination of a constant mean function $\mu_i \in \mathbb{R}$ and the squared exponential function $k_i$
$$k_i(\textbf{x}, \textbf{x}') = \sigma^2\exp\left(-\frac{(\textbf{x}-\textbf{x}')^T(\textbf{x}-\textbf{x}')}{2l_i^2}\right)$$
The kernel matrix $K_i$ is then defined as
$$\textbf{K}_i:=\begin{bmatrix}
k_i(\textbf{x}_1, \textbf{x}_1) & k_i(\textbf{x}_1, \textbf{x}_2) & \dots & k_i(\textbf{x}_1, \textbf{x}_n) \\
\vdots & \vdots & \vdots & \vdots\\
k_i(\textbf{x}_n, \textbf{x}_1) & k_i(\textbf{x}_n, \textbf{x}_2) & \dots & k_i(\textbf{x}_n, \textbf{x}_n)\\
\end{bmatrix}$$
and with abuse of notation define $\textbf{K}=\textbf{K}+\lambda^2I$ (to account for noise in the function evaluations). The posterior distribution on the function $f_i(\textbf{x})$ at any input $\textbf{x}\in\mathcal{X}$~\cite{Rasmussen2005} is given by
$$P(f_i(\textbf{x})|\textbf{x}^{1:n}, f_i^{1:n}) = \mathcal{N}(\mu^{(n)}_i(\textbf{x}), \sigma_i^{(n)}(\textbf{x}) + \lambda^2), ~~ ~ i\in\{1,2,\ldots,M\} \mbox{~where~}$$
\begin{align*}
  \mu^{(n)}_i(\textbf{x}) &= \mu^{(0)} + \textbf{k}_i^T\textbf{K}_i^{-1}(f_i^{1:n} - \mu_i(\textbf{x}^{1:n})) \\ 
  \sigma_i^{(n)}(\textbf{x}) &= k_i(\textbf{x}, \textbf{x})-\textbf{k}_i^T\textbf{K}_i^{-1}\textbf{k}_i \\
  \textbf{k}_i &= \left[k_i(\textbf{x}, \textbf{x}_1) \hspace{5pt}\ldots\hspace{5pt} k_i(\textbf{x}, \textbf{x}_n)\right].
\end{align*}
\begin{algorithm}\label{alg:ei}
    \centering
    \caption{cEI: Composite BO using EI based acquisition function} \label{alg:all_ei}
    \begin{algorithmic}[1]
        \Require $T \xleftarrow{}$ Budget of iterations
        \Require $h(\cdot), f_1(\cdot), \ldots, f_M(\cdot) \xleftarrow{}$ composition and member functions
        \Require $\textbf{X} = \{\textbf{x}_1, \ldots, \textbf{x}_s\} \xleftarrow{} s$ starting points
        \Require $\textbf{F} = \{(f_1(\textbf{x}), \ldots, f_M(\textbf{x}))\}_{\textbf{x} \in \textbf{X}} \xleftarrow{}$  function evaluations at starting points
        \For{$n = s + 1, \ldots, s + T$}
            \For{$i = 1, \ldots, M$}
                \State Fit model $\mathcal{GP}(\mu_i^{(n)}(\cdot), K_i^{(n)}(\cdot, \cdot))$ using evaluations of $f_i$ at points in $\textbf{X}$
            \EndFor
            \State Find new point $\textbf{x}_n$ by optimizing $\text{cEI}(\textbf{x}, L)$ (defined below) 
            \State Get $(f_1(\textbf{x}_n), \ldots, f_M(\textbf{x}_n))$
            \State Augment the data $(f_1(\textbf{x}_n), \ldots, f_M(\textbf{x}_n))$ into $\textbf{F}$ and update $\textbf{X}$ with $\textbf{x}_n$
        \EndFor
        \Function{cEI}{$\textbf{x}, L$}
            \For{$l = 1, \ldots, L$}
                \State Draw $M$ samples $Z_{(l)} \sim \mathcal{N}_M(0_M, I_M)$
                \State Compute $\alpha^{(l)} := \{h(\{\mu^{(n)}_i(\textbf{x}) + \sigma_i^{(n)}(\textbf{x})Z^{(l)}_i\}) - g^*_n\}^+$ 
            \EndFor
            \State \textbf{return} $E_n(\textbf{x})$ = $\frac{1}{L}\Sigma_{l=1}^L\alpha^{(l)}$
        \EndFunction
     \end{algorithmic}
\end{algorithm}
\subsection{cEI and cUCB Acquisition Functions}
For any fixed point $\textbf{x}\in\mathcal{X}$, we use the information about the composition function $h$ to estimate $g$ by first estimating the value of each member function at $\textbf{x}$. However, this is not a straightforward task and needs to be performed in a way similar to the vanilla EI acquisition using Monte Carlo sampling. We propose to use the following acquisition function, that we call as cEI.
\begin{equation}\label{eq:ei_compute}
    E_n(\textbf{x}) = \mathbb{E}_n\left[{h(\{\mu^{(n)}_i(\textbf{x}) + \sigma^{(n)}_i(\textbf{x})Z_i\})} - g_n^*\right]^+  
\end{equation}
where $Z$ is drawn from an $M$-variate normal distribution and $g_n^*$ is the best value observed so far. This acquisition function is similar to EI as we subtract the best observation, $g^*$, so far and only consider negative terms to be 0.  Assuming independent GPs over the functions allows constant time computation of the variance at $\textbf{x}$. However, since each function $f_i$ is being considered an independent variable with mean $\mu_i^{(n)}(\cdot)$ and variance $\sigma_i^{(n)}(\cdot)$, the calculation of $E_n(\boldsymbol{x})$ does not have a closed form and thus, the expectation needs to be evaluated empirically with sampling. Algorithm~\ref{alg:all_ei} provides the complete procedure for doing BO with this acquisition function.

To alleviate this complexity in estimating the acquisition function, we propose a novel UCB-style acquisition function. This function estimates the value of each member function using the GP priors over them and controls the exploration and exploitation factor with the help of the hyperparameter $\lambda_n$: 
\begin{equation}\label{eq:ucb_compute}
   U_n(\textbf{x}) = h(\{\mu^{(n)}_i(\textbf{x}) + \beta_n \sigma^{(n)}_i(\textbf{x})\}) 
\end{equation}
Algorithm~\ref{alg:all_ucb} gives the complete details for using this acquisition function. The user typically starts with a high value for $\beta$ to promote exploration and reduces iteratively to exploit the low reget regions it found. For our experiments, we start with $\beta=1$ and exponentially decay it in each iteration by a factor of 0.99. 
\begin{algorithm}\label{alg:ei}
    \centering
    \caption{cUCB: Composite BO using UCB based acquisition function} \label{alg:all_ucb}
    \begin{algorithmic}[1]
        \Require $T \xleftarrow{}$ Budget of iterations
        \Require $h(\cdot), f_1(\cdot), \ldots, f_M(\cdot) \xleftarrow{}$ composition and member functions
        \Require $\textbf{X} = \{\textbf{x}_1, \ldots, \textbf{x}_s\} \xleftarrow{} s$ starting points
        \Require $\textbf{F} = \{(f_1(\textbf{x}), \ldots, f_M(\textbf{x}))\}_{\textbf{x} \in \textbf{X}} \xleftarrow{}$  function evaluations at starting points
        \Require $\beta\xleftarrow{}$ Exploration factor
        \For{$n = s + 1, \ldots, s + T$}
            \For{$i = 1, \ldots, M$}
                \State Fit model $\mathcal{GP}(\mu_i^{(n)}(\cdot), K_i^{(n)}(\cdot, \cdot))$ using evaluations of $f_i$ at points in $\textbf{X}$
            \EndFor
            \State Find new point $\textbf{x}_n$ suggested by the composition function using Eq.~\ref{eq:ucb_compute}
            \State Get $(f_1(\textbf{x}_n), \ldots, f_M(\textbf{x}_n))$
            \State Augment the data $(f_1(\textbf{x}_n), \ldots, f_M(\textbf{x}_n))$ into $\textbf{F}$ and update $\textbf{X}$ with $\textbf{x}_n$
            \State Update $\beta$
        \EndFor
     \end{algorithmic}
\end{algorithm}

\section{Experiments and Results}\label{sec:experiments}
In this section, we compare the results of our cUCB and cEI algorithms with Vanilla EI, Vanilla UCB and the state-of-the-art BO for Composite Functions (BO-CF)~\cite{https://doi.org/10.48550/arxiv.1906.01537} using HOGP~\cite{maddox2021bayesian} in terms of loss in regret and runtime of the algorithms. We first compare our methods on 3 test functions and then move on to show their applications to three different pricing scenarios. Our code is available \href{https://github.com/kjain1810/Bayesian-Optimization-for-Function-Compositions-with-Applications-to-Dynamic-Pricing}{here}.

Our algorithms are implemented with the help of the BoTorch framework~\cite{botorch} and use the APIs provided by them to declare and fit the GP models. We assume noiseless observations for our results in this section, and the same results can be obtained when we add Gaussian noise to the problem with a fixed mean and variance. We start with 10 initial random points and run our BO algorithms for 70 iterations. We use a system with 96 Intel Xeon Gold 6226R CPU @2.90GHz and 96GB of memory shared between the CPUs.

We compare the performance of different algorithms based on the log of mean minimum regret till each iteration, averaged over 100 runs. In a single BO run, the regret at iteration $i$ in the $k^{th}$ run is defined as $l_i^k = g^* - g(\textbf{x}_i)$ where $g^*$ is the global maximum of the objective function. The minimum regret at iteration $i$ in the $k^{th}$ run is defined as $m_i^k = \min_{1 \leq j \leq i}l_j^k$ and the final metric at iteration $i$ averaged across 100 runs is calculated as 
$$r_i = \log_{10}\left(\frac{1}{100}\sum_{k=1}^{100}m_i^k\right).$$

\subsection{Results on Test Functions}
\begin{figure}[t]
\centering
\begin{subfigure}{.5\textwidth}
  \centering
  \includegraphics[width=0.95\linewidth]{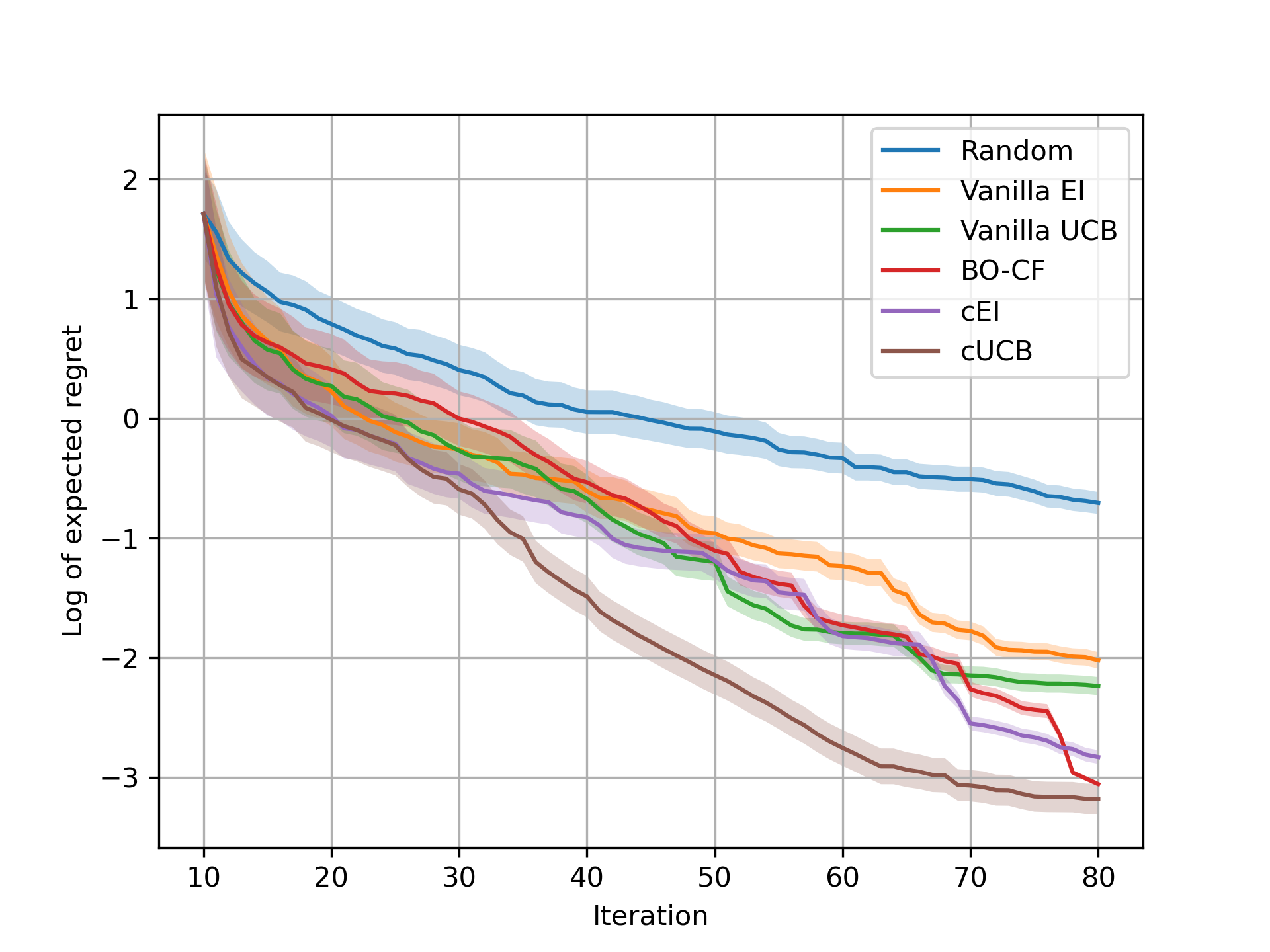}
  \caption{Langermann function}
  \label{fig:langermann}
\end{subfigure}%
\begin{subfigure}{.5\textwidth}
  \centering
  \includegraphics[width=0.95\linewidth]{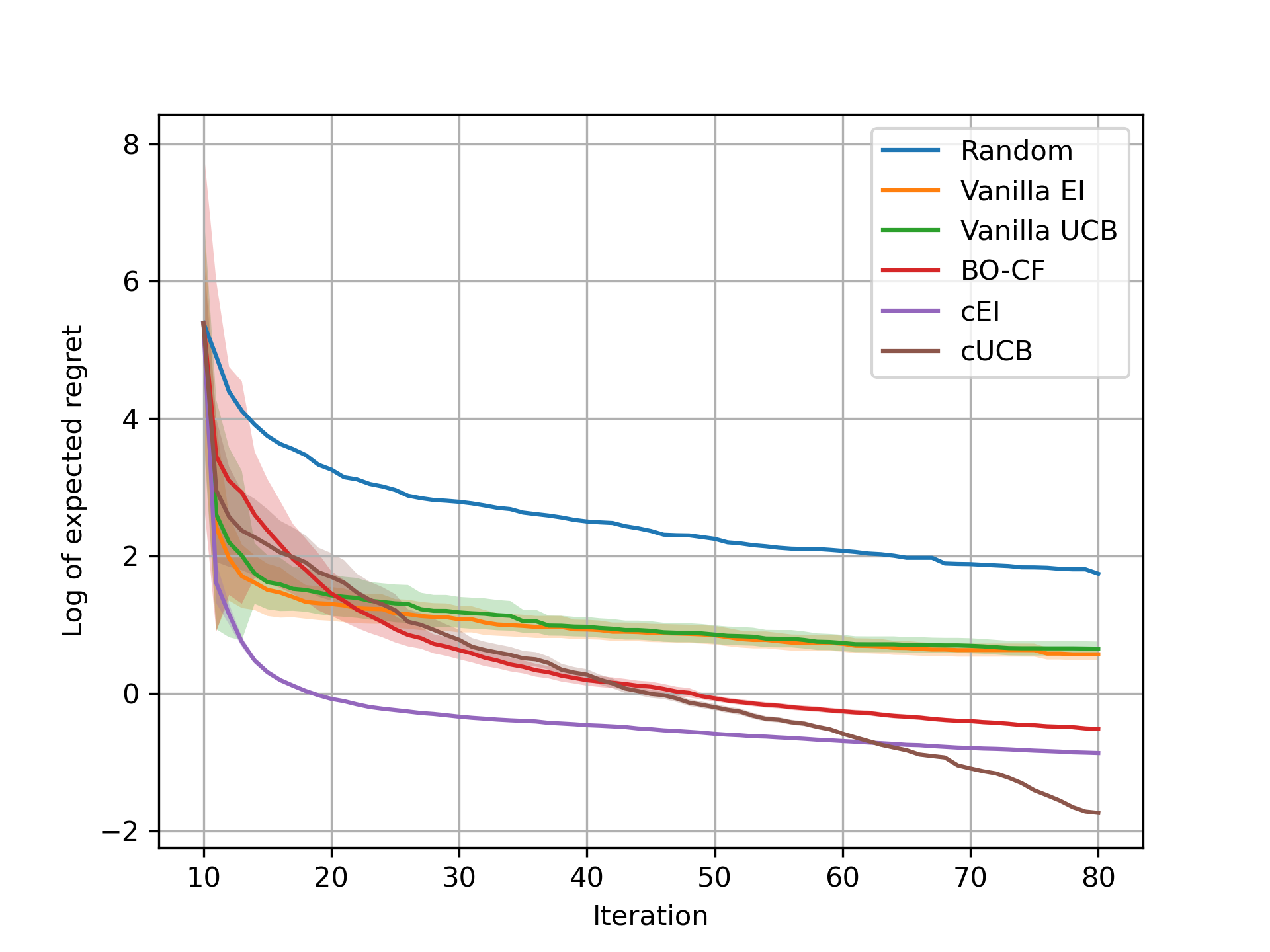}
  \caption{Dixon-Price function}
  \label{fig:dixonprice}
\end{subfigure}
\caption{Log regret for test functions with a composite nature}
\label{fig:testfunctions}
\end{figure}

We first asses our algorithms on three standard test functions~\cite{sfu}:
\subsubsection{Langermann Function:} 
To express this function as a function composition, we consider each outer iteration of the Langermann function to be a separate constituent function, that is, 
$$f_i(\textbf{x}) = \exp\left(-\frac{1}{\pi}\sum\limits_{j=1}^d(x_j-A_{ij})^2\right)\cos\left(\pi\sum\limits_{j=1}^d(x_j-A_{ij})\right).$$
The composition for this will be $h(\{f_i(\textbf{x})\}) = \sum_{i=1}^mc_if_i(\textbf{x})$ with $d=2$, $c=(1,2,5,2,3)$, $m=5$, $A=((3,5),(5,2),(2,1),(1,4),(7,9))$ and domain $\mathcal{X}=[0, 10]^2$. Note that the terms differ only in the columns of hyperparameter $A$ for different member functions and thus, should have a high covariance.

\subsubsection{Dixon-Price Function} In this function, we take the term associated with each dimension of the input to be a separate constituent, that is,
$$f_1(\textbf{x}) = (x_1 - 1)^2, f_i(\textbf{x}) = i(2x_i^2-x_{i-1})^2$$
The composition for this function will be $h(\{f_i(\textbf{x})\}) = \sum_{i=1}^df_i(\textbf{x})$ with $d=5$ and domain $\mathcal{X}=[-10, 10]^d$. Since only consecutive terms in this function share one variable, the member functions do have a non-zero covariance but it will not be as high as in the Langermann function.

\subsubsection{Ackley Function} Here, we build a more complex composition function by considering the terms in the exponents as the member functions, resulting in
$$f_1(\textbf{x})=\sqrt{\frac{1}{d}\sum\limits_{i=1}^dx_i^2} \text{ and } f_2(\textbf{x})=\frac{1}{d}\sum\limits_{i=1}^d\cos(cx_i)$$
$$h(\textbf{x}) = -a\exp(-bf_1(\textbf{x}))-\exp(f_2(\textbf{x}))+a+\exp(1)$$
with $d=5$, $a=20$, $b-0.2$, $c=2\pi$ and domain $\mathcal{X}=[-32.768, 32.768]^d$.

\subsubsection{Results} 
\begin{figure}[t]
\centering
\begin{subfigure}{.5\textwidth}
  \centering
  \includegraphics[width=0.95\linewidth]{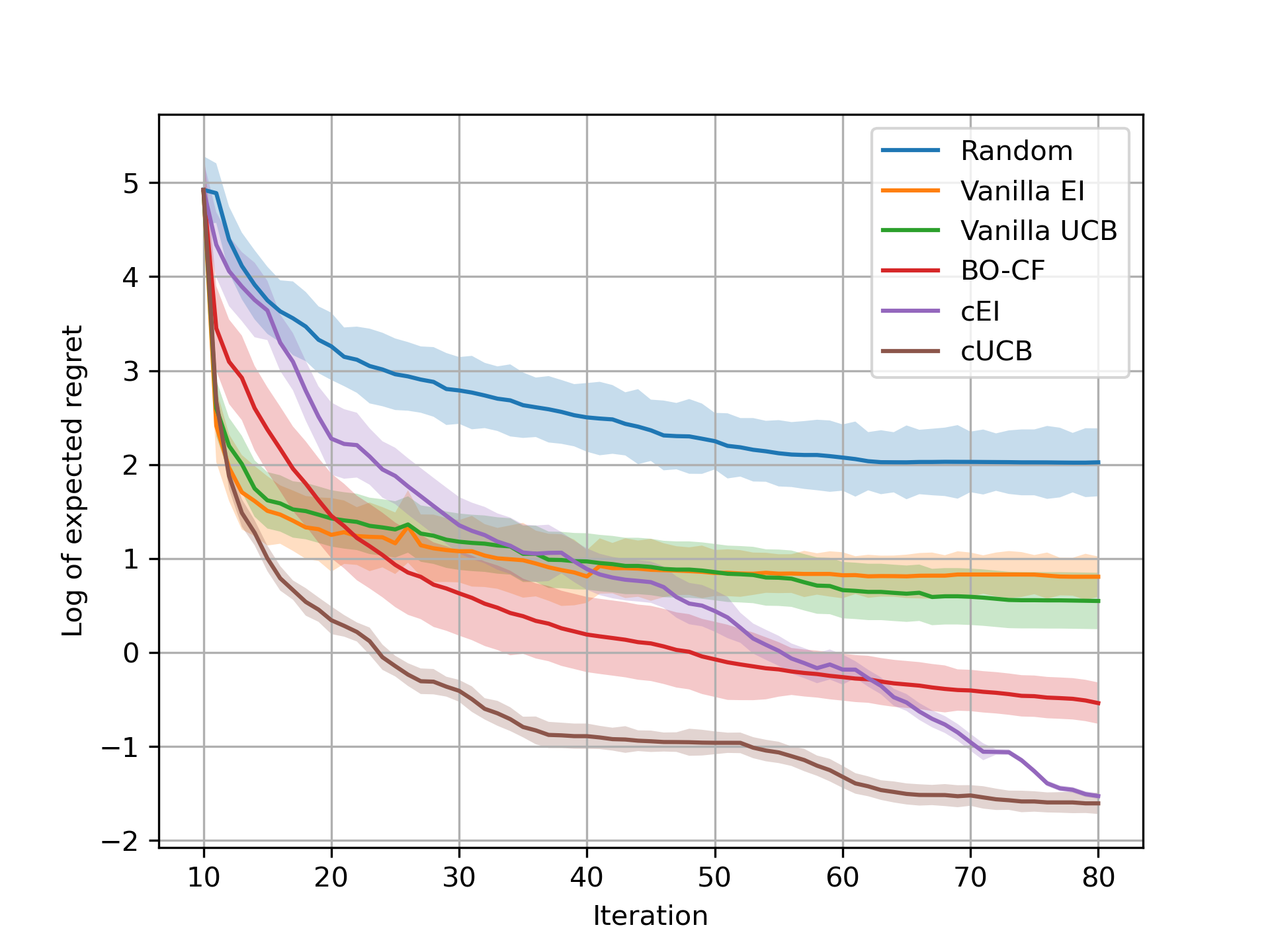}
  \caption{Ackley function}
  \label{fig:regret_ackely}
\end{subfigure}%
\begin{subfigure}{.5\textwidth}
  \centering
  \includegraphics[width=0.95\linewidth]{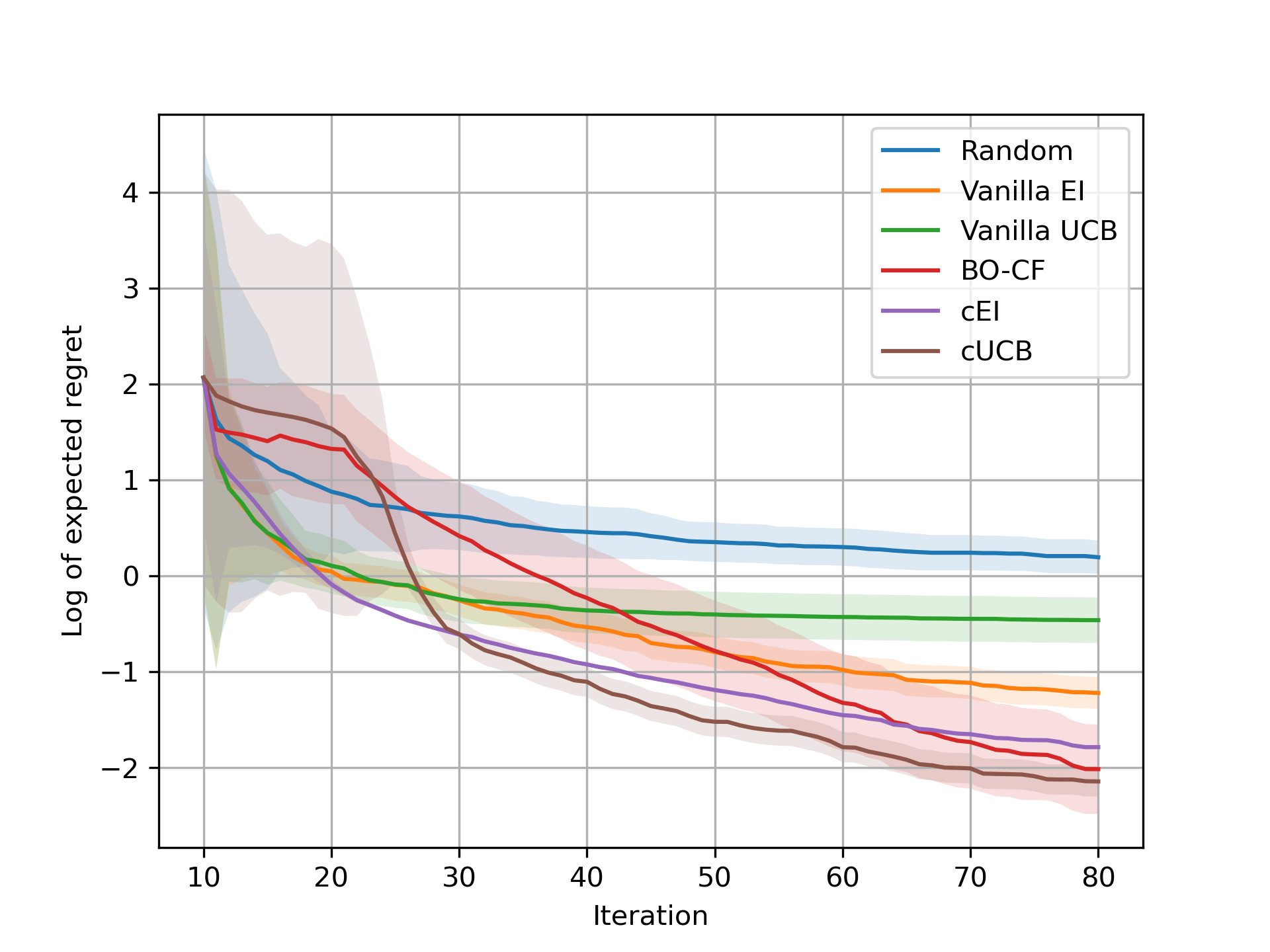}
  \caption{Independent demand model}
  \label{fig:pricing1}
\end{subfigure}
\caption{Log of expected regret for pricing tasks}
\label{fig:testfunctions}
\end{figure}

Figure~\ref{fig:langermann},~\ref{fig:dixonprice} and~\ref{fig:regret_ackely} compares the results of different algorithms. Vanilla EI and UCB algorithms do not consider the composite nature of the function while BO-CF and our methods use the composition defined above. Even with the high covariance between the members in Langermann function, cUCB outperforms BO-CF while the cEI algorithm has a similar performance level. However, when that covariance reduces in the Dixon-Price function, the cEI algorithm performs better than BO-CF while the cUCB algorithm significantly outperforms it. Figure~\ref{fig:regret_ackely} shows that our algorithms work well with complicated composition functions as well and both, cUCB and cEI, outperform BO-CF.

\subsection{Results for Demand Pricing Experiments}

We now test our approach on the demand models discussed in Section~\ref{sec:demand_model}.
\subsubsection{Independent Demand Model:} Recall that in this model, we allow the price at each location to be different and model the demand in a region to depend only on the price therein. We consider 4 regions where each region has a parametric demand functions and randomly chosen parameters. Particularly, we assume that two regions have  a logit demand function ($d(p) = \frac{e^{-z_1-z_2p}}{1+e^{-z_1-z_2p}}$) with $z_1\in\left[1.0, 2.0\right]$, $z_2\in\left[-1.0. 1.0\right]$ and the other two regions have a linear demand function ($d(p) = z_1-z_2p$) with $z_1\in\left[0.75, 1\right]$ and $z_2\in[2/3, 0.75]$. The domain for this model is $\mathcal{X}=[0,1]^4$ and the composition is 
The composition function looks $h(d_1(\textbf{p}), d_2(\textbf{p}), d_3(\textbf{p}), d_4(\textbf{p})) = \sum_{i=1}^4p_id_i(p_i)$.
%  like Equation(\ref{eq:firstdemand}) and domain space is $\mathcal{X}=[0,1]^4$
% \begin{equation}\label{eq:firstdemand}
%     h(d_1(\textbf{p}), d_2(\textbf{p}), d_3(\textbf{p}), d_4(\textbf{p})) = \sum\limits_{i=1}^4p_id_i(p_i)
% \end{equation}
\subsubsection{Correlated Demand Model:} In this example, we assume that different products are for sale at different prices, and there is a certain correlation between demand for different products via their prices. We consider the case of 2 products where one of the product has a demand function governed by the Matyas function~\cite{sfu}. We assume that the demand for the second product is governed by the Booth function~\cite{sfu}. More specifically, the function composition and constituent functions are as below, where the domain for the problem is $\mathcal{X}=[0,10]^2$:
\begin{align*}
    d_1(\textbf{p}) &= 8(100 - \text{Matyas}(\textbf{p})) \\
    d_2(\textbf{p}) &= 1154 - \text{Booth}(\textbf{p})
\end{align*}
where Matyas function is defined as $\text{Matyas}(\textbf{x}) = 0.26(x_1^2+x_2^2)-0.48p_1p_2$. Similarly, Booth function is defined as $\text{Booth}(\textbf{x}) = (x_1 + 2x_2 -7)^2 + (2x_1+x_2-5)^2$. The composition for this will be $h(d_1(\textbf{p}), d_2(\textbf{p})) = p_1d_1(\textbf{p}) + p_2d_2(\textbf{p})$.
\subsubsection{Identical price Model:} In this example, we assume that a commodity is sold for same price at two different regions but the willingness to pay variable for customers in the two regions is different. We assume that the willingness to pay distribution in one region follows  exponential distribution with $\lambda = 5.0$. n the other region, this is assumed to be a gamma distribution with $\alpha=10.0, \beta=10.0$ The resulting function composition is given by :
$h(d_1(p), d_2(p)) = pd_1(p) + pd_2(p)$.
\subsubsection{Results} 
\begin{figure}[t]
\centering
\begin{subfigure}{.5\textwidth}
  \centering
  \includegraphics[width=0.95\linewidth]{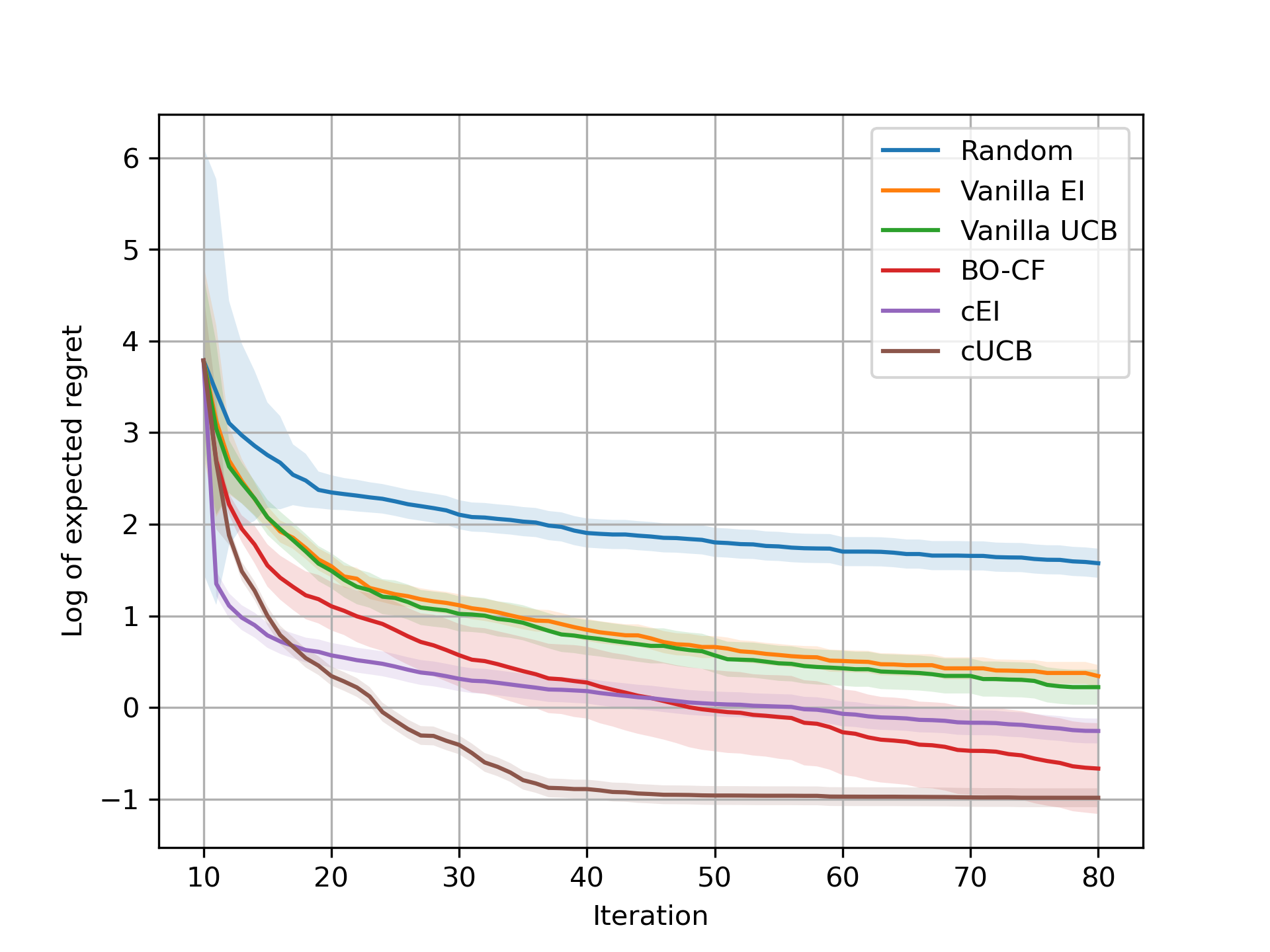}
  \caption{Correlated demand model}
  \label{fig:pricing2}
\end{subfigure}%
\begin{subfigure}{.5\textwidth}
  \centering
  \includegraphics[width=0.95\linewidth]{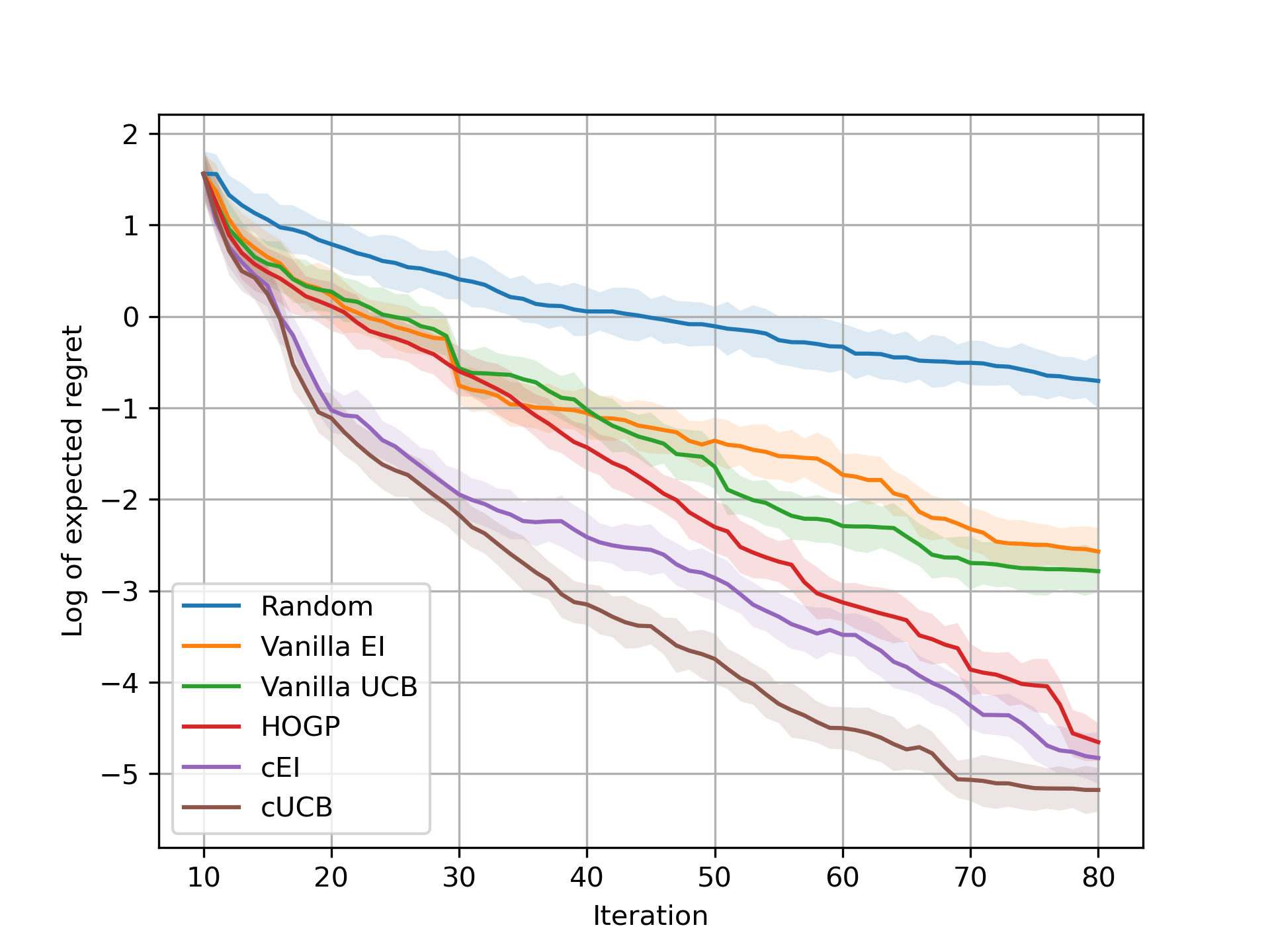}
  \caption{Identical price model}
  \label{fig:pricing_3}
\end{subfigure}
\caption{Log of expected regret for pricing tasks}
\label{fig:testfunctions}
\end{figure}

Figures~\ref{fig:pricing1}, \ref{fig:pricing2} and \ref{fig:pricing_3} compare the results of these dynamic pricing models for the different BO algorithms. 
Our algorithms perform well even in higher dimensions of input and member functions with cUCB marginally outperforming BO-CF in the first model. cUCB matches the minimum regret in the second model and converges to it much faster than BO-CF. In the case of the third model, having independent GP's performs better than BO-CF with cEI.
\begin{table}[t]
\centering
\begin{tabular}{lccccccc}
    \hline
    Task & EI & UCB & cEI & cUCB & HOGP\\
    \hline
    Langermann function & 1.74 & 1.73 & 9.71 & 9.34 & 36.30 \\
    Dixon-Price function & 1.70 & 1.69 & 14.47 & 12.71 & 41.72 \\
    Ackley function & 1.63 & 1.55 & 11.56 & 11.37 & 47.91 \\
    Independent demand model & 2.05 & 2.03 & 7.85 & 7.12 & 19.12 \\
    Correlated demand model & 3.91 & 3.29 & 9.63 & 9.19 & 53.01 \\
    Identical price model & 2.55 & 2.38 & 7.27 & 7.81 & 24.37 \\
    \hline
\end{tabular}
\caption{Run-time for 70 iterations across algorithms in seconds}
\label{tab:runtimes}
\end{table}
\subsection{Runtime Comparisons with State of the Art}
Along with the performance of our algorithms being superior in terms of regret, the methodology of training independent GPs is significantly faster in terms of run time. As shown in Table~\ref{tab:runtimes}, both of our algorithms are between 3 to 4 times faster than BO-CF using HOGP on average and their run time increases linearly with the number of member functions in the composition when compared to vanilla EI and UCB.
By not having to compute the inverse matrix for estimating the lower Cholesky factor of the covariance matrix, we gain large improvements in run time.
The elimination of inverse matrix computation 
while estimating with the help of lower Cholesky factor of the covariance matrix results in the large improvement in run time over BO-CF. Also note that our UCB variant is marginally faster than the EI variant as well due to the elimination of MC sampling in the process.

\section{Conclusion}\label{sec:future_works}
In this work, we have proposed EI and UCB based BO algorithms, namely cEI and cUCB for optimizing functions with a composite nature. We further apply our algorithms to the revenue maximization problem and test our methods on different market scenarios. We show that our algorithms, particularly cUCB, outperforms vanilla BO as well as the current state of the art BO-CF algorithm. Our algorithms are  computationally superior because they do not require multiple Cholesky decompositions as required in the BO-CF algorithm.

As part of future work, we would like to provide  theoretical bounds on cumulative regret for the proposed algorithms. We would also like to see the applicability of the proposed algorithms in hyper-parameter tuning for optimizing F1 score. It would also be interesting to propose BO algorithms for an extended model wherein the member functions can be probed independently from each other at different costs.

% \textcolor{blue}{Conclude by saying we outperform classic acquistion functions in regret and eicf/hogp in runtime. Thus, we propose a good balance. Think of some future works maybe or change the section title.}

% \textbf{expand on the future work part .... say theoretical guarantees on regret is one future work. Second future work is to study the problem of precision and recall. The other is to study the case when one has to evaluate different functions independently and when there is a different cost for evaluating different functions, then one has to trade off different function evaluations.}

% \textcolor{red}{Loss in revenue due to regret in comparison to eicf can also be accounted for in savings of hardware requirements for our methodology maybe?}

% \textcolor{blue}{Future work: look at applications of causal graph BO to assist in supply chain optimization. this lays the groundwork for that.}

% \textcolor{red}{fix references, they are not consistent in the information they contain.... somewhere you have doi, reference 5 is going to 5 lines .... reference 1 and 2 have DOI given twoice ,,, arxiv link given twice .. this is enough ground for rejection by virtue of being shabby}

\bibliographystyle{splncs04}
\bibliography{myrefs}
\end{document}